\documentclass[letterpaper, 10 pt, conference]{ieeeconf}  

\IEEEoverridecommandlockouts                              
\usepackage[utf8]{inputenc}
\usepackage[T1]{fontenc}
\usepackage{graphicx} 
\usepackage{cite}
\usepackage{url}
\usepackage{booktabs}
\usepackage{tikz}
\usepackage{amsmath}
\usepackage{amssymb}
\usetikzlibrary{shapes.geometric, arrows}
\usepackage{mathrsfs}
\usepackage{hyphenat}
\usepackage{algorithm}
\usepackage{tabularx}
\usepackage{colortbl}
\usepackage{algpseudocode}
\usepackage{hhline}
\usepackage{placeins}

\title{\LARGE \bf
Experimental Analysis of Quadcopter Drone Hover Constraints for Localization Improvements
}

\author{Uthman Olawoye$^{1}$, David Akhihiero$^{1}$ and Jason N. Gross$^{1}$
\thanks{$^{1}$Department of Mechanical, Materials and Aerospace Engineering, West Virginia University}
}

\begin{document}

\maketitle
\thispagestyle{empty}
\pagestyle{empty}

\begin{abstract}

In this work, we evaluate the use of aerial drone hover constraints in a multisensor fusion of ground robot and drone data to improve the localization performance of a drone. In particular, we build upon our prior work on cooperative localization between an aerial drone and ground robot that fuses data from LiDAR, inertial navigation, peer-to-peer ranging, altimeter, and stereo-vision and evaluate the incorporation knowledge from the autopilot regarding when the drone is hovering.  This control command data is leveraged to add constraints on the velocity state.  Hover constraints can be considered important dynamic model information, such as the exploitation of zero-velocity updates in pedestrian navigation. We analyze the benefits of these constraints using an incremental factor graph optimization.  Experimental data collected in a motion capture faculty is used to provide performance insights and assess the benefits of hover constraints.

\end{abstract}

\section{Introduction}
Localization constraint equations, such as zero-velocity constraints, are powerful in low-cost inertial navigation applications as a method to reduce the rate of dead-reckoning drift.  Most prominently, pedestrian navigation applications have leveraged the natural walking gait to regularly use zero-velocity information and reduce the rate of drift of inertial navigation ~\cite{abdulrahim2014understanding, park2010zero, foxlin2005pedestrian}. In 2020, Wahlstrom et al. \cite{wahlstrom2020fifteen} recounted the developments in the use of zero-velocity updates in foot-mounted inertial navigation and summarized open questions related to sensors, algorithms, and the need for improvements in evaluation methods.

Zero-velocity constraints have also proven powerful for wheeled mobile robots.  For example, Kilic et al. \cite{kilic2019improved} showed that properly leveraging zero-velocity constraints during the periodic stopping of a wheeled mobile robot along with other available constraints such as nonholonomic and zero angular rate constraints allowed low-cost dead-reckoning to lead to significantly lower drift rates.  An error reduction of more than ten times was demonstrated in several test cases compared to not using zero-velocity constraints.  These approaches on wheeled platforms were extended to consider autonomously triggering a stop condition to reduce drift based on terrain conditions~\cite{kilic2021slip} and for cooperative robot application~\cite{kilic2023evaluation}.  Although it is feasible to reasonably ensure that a wheeled robot can be stopped to leverage zero-velocity constraints, this is not possible for aerial platforms.

For aerial drone localization, researchers have evaluated the potential constraining localization drift with the use of a vehicle dynamic model.  For example, Khaghani and Skaloud \cite{khaghani2016autonomous} demonstrated the advantages of Vehicle Dynamic Model (VDM) aided inertial navigation to reduce solution drift during Global Navigation Satellite System (GNSS) outages using a Monte-Carlo simulation.  While valuable, accurate VDM requires extensive parameter identification and is subject to model parameter errors.

This paper extends previous work that evaluated aerial drone multisensor localization approaches relative to a cooperative ground robot~\cite{akhihiero2024cooperative} by incorporating zero-velocity-like constraints in a factor graph \cite{dellaert2017factor} when the drone autopilot indicates a hovering condition. The approach of using a hovering condition constraint was incorporated in our earlier work that formulated a similar solution with an error-state Extended Kalman Filter, however, it use was not rigorously analyzed and assessed~\cite{gross2019field}. Further, it's value is anticipated to be greater in a factor graph that iterates over the solutions trajectory over time. This additional constraint aims to improve localization accuracy by reducing drift in velocity estimates during hovering periods.  While based on a heuristic that assumes the onboard autopilot correctly identifies a hovering condition that could be wrong, we explore the potential of leveraging hovering conditions as they can be considered a reduced complexity VDM for special conditions.  Further, while hovering is not strictly a zero-velocity condition, we explore tuning the uncertainty of the constraint and assess its impact. We leverage the same experimental setup and recorded data as our recent paper ~\cite{akhihiero2024cooperative} and focus this conference paper on assessing only the impact of this additional information from hovering.  The next section shares the general formulation adopted.  This is followed by a discussion of the experimental setup and a discussion of the localization performance results. The paper ends with some concluding remarks and ideas for future work.


\section{Formulation}
Our previous work \cite{akhihiero2024cooperative} formulated the UAV state estimation problem as a factor graph optimization (FGO) \cite{dellaert2012factor, das2021review} that incrementally smooths the state trajectory. Factor graphs have been shown to be able to solve nonlinear systems more accurately than Kalman filters and are a desirable choice for GNSS degraded/denied environments \cite{taylor2024factor}. The state vector $X_i$ at each time step $i$ comprised the UAV's 3D pose, velocity, and IMU biases. The optimization framework incorporated multiple measurement factors:
\begin{itemize}
    \item Prior factors ($\Psi_{\text{prior}}$) from visual odometry and LiDAR segmentation
    \item Elevation factors ($\Psi_{\text{elevation}}$) from altimeter measurements
    \item Range factors ($\Psi_{\text{range}}$) between UAV and UGV using UWB measurements
    \item IMU factors between consecutive states using pre-integrated measurements
\end{itemize}
This formulation achieved robust state estimation by fusing inertial propagation with visual and range updates through maximum a posteriori (MAP) estimation.

In this updated formulation, we introduce a unary factor we call \textbf{hovering factor} (\(\Psi_{hover}\)) that enforces a zero-velocity constraint when the UAV is detected to be in hover mode. The autopilot detects hover mode, and a new unary factor is added at the corresponding state node, setting the velocity to zero.
The hover factor is defined as:

\begin{equation}
    \Psi_{\text{hover}}(X_i) = 
        \|v_i\|_{\Sigma_{h}}^2
\end{equation}
Thus, the enhanced optimization objective becomes:
\begin{equation}
    \begin{split}
    X = \underset{X}{\arg \min} \sum_{i} \| X_i -X_0\|_{\Sigma_{p}}^2 + \|Z_i - h(X_i) \|_{\Sigma_{el}}^2 + \ldots\\
    \|Z_i - h(X_i) \|_{\Sigma_{el}}^2 + \| X_i - X_{UGV}\|_{\Sigma_{range}}^2 + \ldots \\
    \sum_{j} \|X_{j} - f(X_{j-1}) \|_{\Sigma_{IMU}}^2 + \sum_{i \in \mathcal{H}} \|v_{i}\|_{\Sigma_{hover}}^2
    \end{split}  
    \label{x_map_optimized}
\end{equation}
where $\mathcal{H}$ denotes the states at which hovering is detected. A visual representation of the formulation is presented in figure \ref{fg} below.

\begin{figure}[htb]
    \centering
    \resizebox{0.48\textwidth}{!}{ 
    \begin{tikzpicture}[node distance=2cm, auto]
    
    \node [above] at (-3,4) {Fast-LIO for the UGV}; 
    
    \draw [dashed, thin, draw=red] (-9,2) -- (3,2);
    
    \node [circle, draw, minimum size=1.4cm] (c1) at (-8,3) {$X_{\text{UGV}}$};
    \node [circle, draw, minimum size=1.4cm] (c2) at (-6,3) {$X_{\text{UGV}}$};
    \node [circle, draw, minimum size=1.4cm] (c3) at (-4,3) {$X_{\text{UGV}}$};
    \node [circle, draw, minimum size=1.4cm] (c4) at (-2,3) {$X_{\text{UGV}}$};
    \node [circle, draw, minimum size=1.4cm] (c5) at (0,3) {$X_{\text{UGV}}$};
    \node [circle, draw, minimum size=1.4cm] (c6) at (2,3) {$X_{\text{UGV}}$};
    
    \draw [dashed] (c1) -- (c2);
    \draw [dashed] (c2) -- (c3);
    \draw [dashed] (c3) -- (c4);
    \draw [dashed] (c4) -- (c5);
    \draw [dashed] (c5) -- (c6);
    \draw [dashed] (c6) -- +(2,0) ;
    
    \node[draw, circle, minimum size=1.4cm] (x1) at (-6,-3) {$X_{t0}$};
    \node[draw, rectangle, left of=x1] (f_p1) {$\Psi_{\text{prior}}$};
    \node[draw, rectangle, right of=x1] (f_i1) {$\Psi_{i-1, i}^{\text{preintIMU}}$}; 
    
    \node[draw, circle, minimum size=1.4cm ,right of=f_i1] (x2) {$X_{t1}$};
    \node[draw, rectangle, below of=x2] (f_e1) {$\Psi_{\text{elevation}}$};
    \node[draw, rectangle, left of=f_e1] (f_p2) {$\Psi_{\text{prior}}$};
    \node[draw, rectangle, above of=x2] (f_r1) {$\Psi_{\text{range}}$};
    
    \node[draw, circle, minimum size=1.4cm , above of=f_r1] (x_b1) {$X_{\text{UGV, t1}}$}; 
    \node[draw, rectangle, right of=x2] (f_i2) {$\Psi_{i, i+1}^{\text{preintIMU}}$};
    
    \node[draw, circle, minimum size=1.4cm ,right of=f_i2] (x3) {$X_{t2}$};
    \node[draw, rectangle, below of=x3] (f_e2) {$\Psi_{\text{elevation}}$}; 
    \node[draw, rectangle, above of=x3] (f_r2) {$\Psi_{\text{range}}$};
    \node[draw, rectangle, left of=f_e2] (f_p3) {$\Psi_{\text{prior}}$};
    \node[draw, rectangle, right of=f_e2] (f_h) {$\Psi_{\text{hover}}$};
    
    \node[draw, circle, minimum size=1.4cm ,above of=f_r2] (x_b2) {$X_{\text{UGV, t2}}$};
    \node[draw, rectangle, left of=x_b1] (p_b1) {$\Psi_{\text{UGV,prior}}$};
    \node[draw, rectangle, left of=x_b2] (p_b2) {$\Psi_{\text{UGV,prior}}$};
    
    \draw [dashed] (x3) --  +(2,0);
    \draw [dashed, draw=blue, -latex] (c3) -- (p_b1);
    \draw [dashed, draw=blue, -latex] (c5) -- (p_b2);
    
    \path
    (x1) edge (f_p1)  
    (x1) edge (f_i1)
    (f_i1) edge (x2)
    (x2) edge (f_e1)
    (x2) edge (f_r1)
    (x2) edge (f_p2)   
    (f_i2) edge (x2)
    (x3) edge (f_e2)
    (x3) edge (f_p3)
    (x3) edge (f_h)
    (x3) edge (f_r2)
    (f_i2) edge (x3)
    (f_r1) edge (x_b1)
    (f_r2) edge (x_b2)
    (x_b1) edge (p_b1)
    (x_b2) edge (p_b2);
    
    \end{tikzpicture}
    }
    \caption{GNSS-denied UAV-UGV navigation system incremental FGO.} 
    \label{fg}
\end{figure}
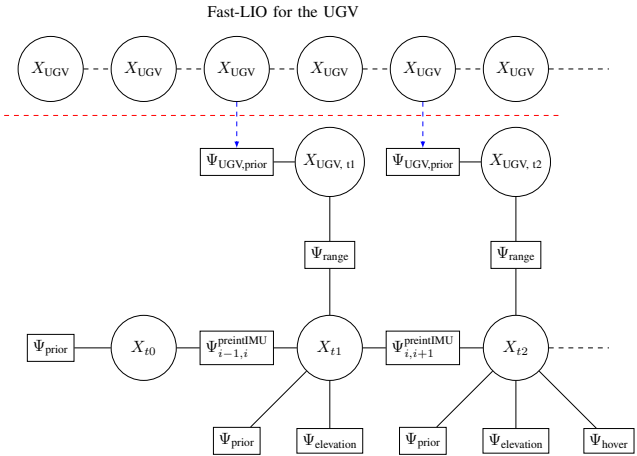

For additional details of the formulation, the reader can find them in the paper that this study leverages and builds upon ~\cite{akhihiero2024cooperative}.

\section{Experimental Set-up}
\begin{figure}[htb]
    \centering
    \includegraphics[width=0.45\textwidth]{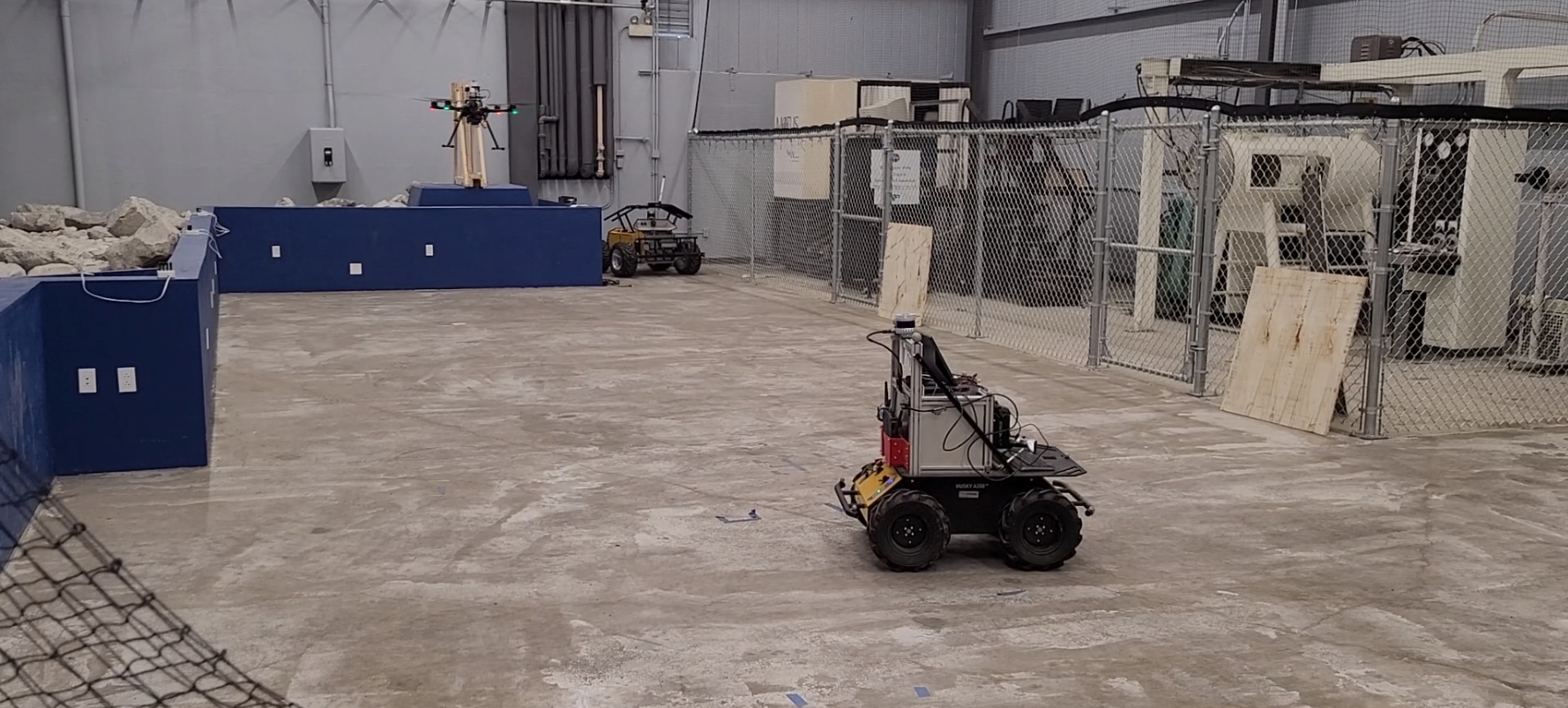}
    \caption{Motion Capture Facility with UAV and UGV Navigating Autonomously During Experimental Test. (WVU Photo)}
    \label{mocap_room}
\end{figure}

Similar to the setup in our previous work \cite{akhihiero2024cooperative}, to evaluate the performance of the estimation algorithm, we conducted a field test involving simultaneous autonomous UAV flights and UGV movements inside a mmotion capture facility shown in Figure \ref{mocap_room} equipped with 30 VICON Vantage V5 cameras. These cameras provided real-time feedback at a high frame rate for precise motion tracking. Both UAV and UGV trajectories were pre-planned, and the UAV’s pose was determined using VICON motion tracking with reflective markers attached to the UAV. In this experiment, the FGO position estimates were not used for closed-loop control as the focus was on evaluating the localization accuracy. The UGV motion was analyzed after the experiment using VICON data and the FAST-LIO algorithm \cite{xu2021fast}. The following noise parameters (sigma/standard deviations) were used for the factor graph.

\begin{table}[h]
    \centering
    \caption{Noise Model Sigmas for Various Factors}
    \begin{tabular}{l l c}
        \toprule
        \textbf{Description} & \textbf{Parameter} & \textbf{Value} \\
        \midrule
        GPS Factor Attitude           & $\Sigma_{\text{GA}}$  & 0.1 m \\
        GPS Factor Position X         & $\Sigma_{\text{GPx}}$ & 0.01 \\
        GPS Factor Position Y         & $\Sigma_{\text{GPy}}$ & 0.2 m \\
        GPS Factor Position Z         & $\Sigma_{\text{GPz}}$ & 0.2 m \\
        IMU Preint Factor Attitude    & $\Sigma_{\text{IPA}}$ & 0.01 rad. \\
        IMU Preint Factor Position    & $\Sigma_{\text{IPP}}$ & 0.35 m\\
        Badger Position Prior         & $\Sigma_{\text{BP}}$  & 0.1 m \\
        Range Factor                  & $\Sigma_{\text{R}}$   & 0.1 m \\
        Elevation Factor              & $\Sigma_{\text{REF}}$ & 0.1 m \\
        Velocity Prior Factor         & $\Sigma_{\text{V}}$   & 0.02 m/s \\
        Hover Factor                  & $\Sigma_{\text{H}}$   & 0.01 m/s (0.05 m/s also tested)\\
        \bottomrule
    \end{tabular}
    \label{tab:noise_model}
\end{table}

The UAV was flown at an altitude of 2 meters above the UGV to avoid collisions with the UGV, while the UGV followed an 8-shaped trajectory. The total duration of the test was approximately 150 seconds and the collected data was analyzed by ROS bag file replay \cite{quigley2009ros}. For a full description of the experimental setup, refer to our previous work \cite{akhihiero2024cooperative}.

\section{Results}
\begin{figure}[htb!]
    \centering
    \includegraphics[width=0.49\textwidth]{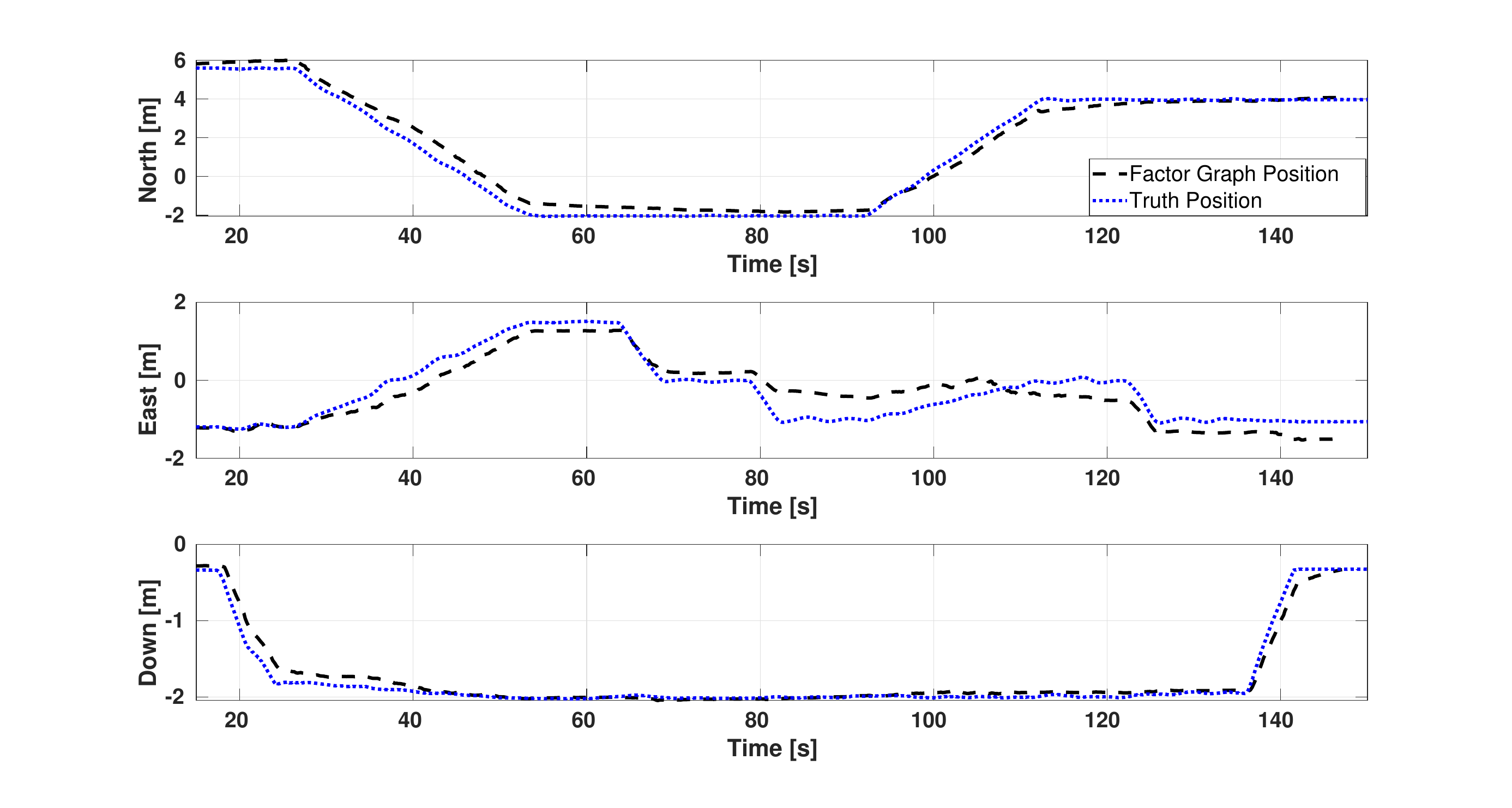}
    \caption{FGO position (with hover constraint) compared with the truth position.}
    \label{fg_pos}
\end{figure}

\begin{figure}[htb]
    \centering
    \includegraphics[width=0.49\textwidth]{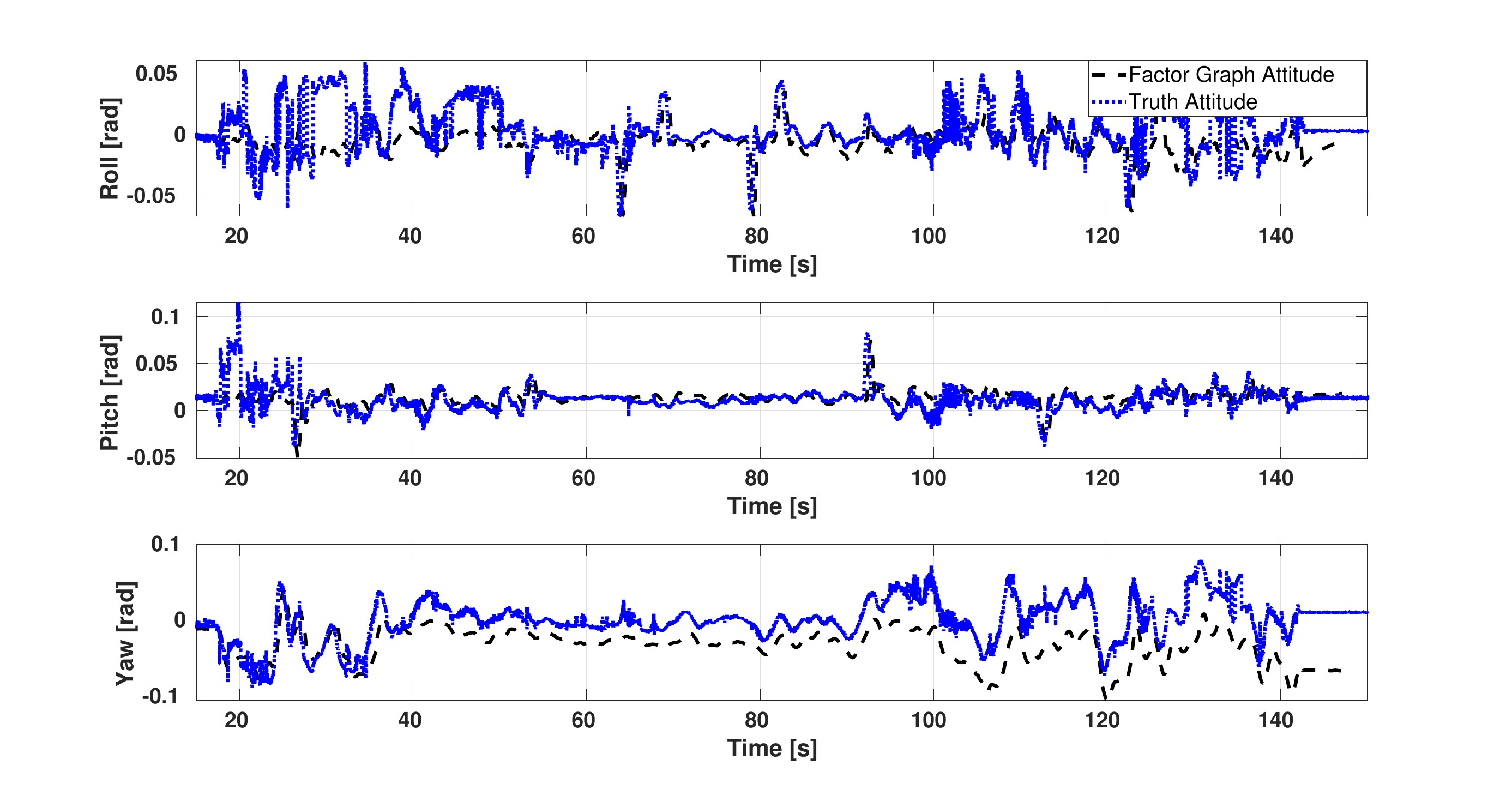}
    \caption{FGO attitude (with hover constraint) compared with the truth attitude.}
    \label{fg_att}
\end{figure}

\begin{figure}[htb]
    \centering
    \includegraphics[width=0.49\textwidth]{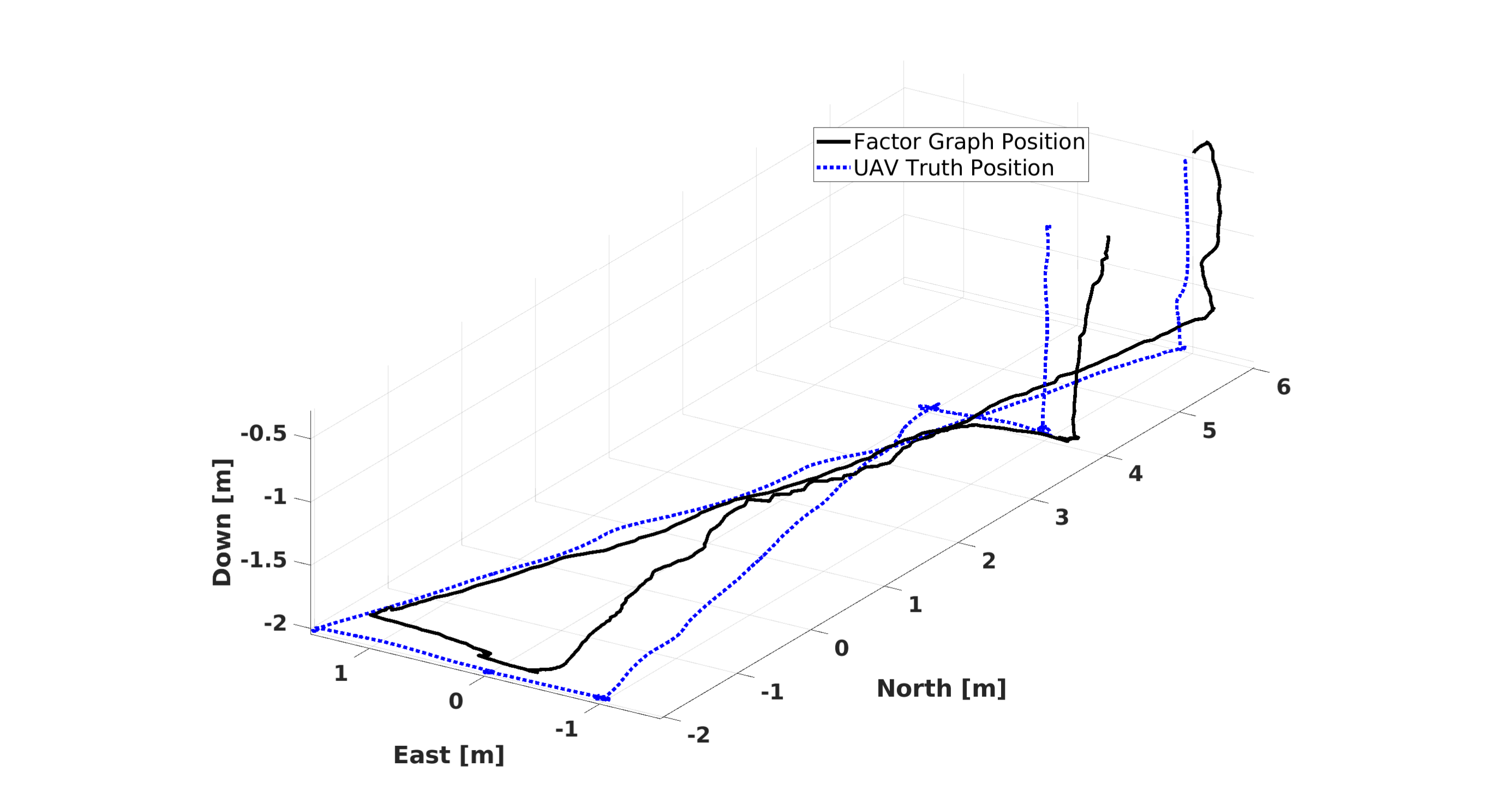}
    \caption{3D plot of FGO position (with hover constraint) compared with the truth position.}
    \label{fg_pos3d}
\end{figure}

The results of our incremental Factor Graph Optimization (FGO) approach are presented in this section. To offer the most insight, an experimental test to evaluate the solution of our estimation algorithm with multiple measurement updates, including the hover constraint update with different noise models ranging from low to high confidence was conducted. The tested measurement models include: (1) LiDAR position only, (2) LiDAR position + UWB range, (3) LiDAR position + UWB range + altimeter, (4) LiDAR position + UWB range + altimeter + hover (evaluated with three different noise levels), and (5) LiDAR position + UWB range + altimeter + hover + velocity. Figures \ref{fg_pos} and \ref{fg_att} show the comparison of our position and attitude estimates with the truth position and attitudes, while Tables \ref{rms_error} and \ref{max_error} present the root mean square (RMS) and maximum errors across all axes of the state estimate. A 3D plot of the truth position compared with the estimated position is shown in Figure \ref{fg_pos3d}.

To assess the value of the hovering constraint, different sensor combinations were selectively included particular sensor inputs in each estimation model during the post-processing of the recorded data to evaluate the effects of various measurement updates. Due to frame transformations, SLAM localization, and sparse LiDAR observations, the initial tests using only inertial navigation and LiDAR position updates revealed significant drift in the state estimate, especially in the east and down directions. Also, when the UAV moved out of the LiDAR's field of view, or the heuristic-based UAV identification was too conservative, tracking failed.

\begin{table*}[h!]
\caption{Factor Graph RMS Error}
\centering
\begin{tabularx}{\textwidth}{|>{\raggedright\arraybackslash}p{4cm}|X|X|X|X|X|X|X|}
\hline
\textbf{Measurement Updates} & \textbf{North (m)} & \textbf{East (m)} & \textbf{Down (m)} & \textbf{3D (m)} & \textbf{Roll (rad)} & \textbf{Pitch (rad)} & \textbf{Yaw (rad)} \\
\hline
\nohyphens{\textbf{IMU + LiDAR Updates}} & 110.685 & 82.665 & 49.138 & 146.626 & 0.030 & 0.027 & 0.054 \\
\hline
\nohyphens{\textbf{IMU + LiDAR + UWB Updates}} & 2.185 & 2.303 & 0.570 & 3.225 & 0.026 & 0.022 & 0.075 \\
\hline
\nohyphens{\textbf{IMU + LiDAR + UWB + Altimeter Updates}} & 2.042 & 2.321 & 0.121 & 3.094 & 0.026 & 0.021 & 0.073 \\
\hline
\nohyphens{\textbf{IMU + LiDAR + UWB + Altimeter + Hover (0.01) Updates}} & 0.451 & 2.272 & 0.067 & 2.318 & 0.019 & 0.011 & 0.036 \\
\hline
\nohyphens{\textbf{IMU + LiDAR + UWB + Altimeter + Hover (0.05) Updates}} & 1.535 & 2.425 & 0.133 & 2.873 & 0.024 & 0.014 & 0.051 \\
\hline
\nohyphens{\textbf{IMU + LiDAR + UWB + Altimeter + Hover (0.1) Updates}} & 0.419 & 2.140 & 0.095 & 2.183 & 0.024 & 0.014 & 0.041 \\
\hline
\nohyphens{\textbf{IMU + LiDAR + UWB + Altimeter + Velocity Updates}} & 0.315 & 0.313 & 0.050 & 0.447 & 0.021 & 0.018 & 0.061 \\
\hline
\nohyphens{\textbf{IMU + LiDAR + UWB + Altimeter + Hover + Velocity Updates}} & 0.404 & 0.362 & 0.091 & 0.550 & 0.022 & 0.012 & 0.039 \\
\hline
\end{tabularx}
\label{rms_error}
\end{table*}

The introduction of UWB updates greatly enhanced localization by lowering the 3D RMS error to about 3.2 m. Further integration of altimeter updates resulted in additional accuracy gains, and the RMS error was reduced to less than 1 m with the addition of velocity updates. We noticed further improvements in our state estimates when we included the hover measurement updates. In recognition of the fact that hovering does not imply exactly zero velocity when using the hover updates, we compared the estimates with three different noise levels in our model: 0.01, 0.05, and 0.1. A noise level of 0.01 implies high confidence in the drone velocity being close to zero and leads to tighter constraints in the graph. This appears to improve the roll, pitch, and yaw estimates, and this intuitively makes sense since the roll and pitch need to be almost zero during hovering to prevent drift. However, the 0.1 noise level showed the lowest 3D rms error and still had some improvement in the attitude error. This noise level allows more velocity variations during hover that may better match the real hover behavior. The noise level of 0.05 is in an in-between region where the constraints are neither tight enough to provide strong correction nor loose enough to allow natural hover motion, and this may lead to inconsistent corrections, making the factor graph struggle to reconcile other sensor measurement updates with hover updates. In this comfiguration, the 3D RMS error is reduced another meter to 2.18 m. Our results also include the case where hover updates are excluded, and velocity updates are included, as well as another where both hover and velocity updates are included. As shown in our previous work \cite{akhihiero2024cooperative}, including velocity updates greatly reduces the errors across all axes. From our results, the inclusion of hover updates with velocity updates appears to slightly increase the rms errors except for the yaw error, which appears to reduce slightly. 

The reason for this could be a conflict between hover updates and velocity updates that cause small inconsistencies in the factor graph. A reduction in the yaw error makes sense as the yaw typically remains more stable during hover. This suggests that hover updates provide limited benefits when velocity updates are already included.

\begin{table*}[h!]
\caption{Factor Graph Max Error}
\centering
\begin{tabularx}{\textwidth}{|>{\raggedright\arraybackslash}p{4cm}|X|X|X|X|X|X|X|}
\hline
\textbf{Measurement Updates} & \textbf{North (m)} & \textbf{East (m)} & \textbf{Down (m)} & \textbf{3D (m)} & \textbf{Roll (rad)} & \textbf{Pitch (rad)} & \textbf{Yaw (rad)} \\
\hline
\nohyphens{\textbf{IMU + LiDAR Updates}} & 589.978 & 412.305 & 375.307 & 722.659 & 0.087 & 0.071 & 0.108 \\
\hline
\nohyphens{\textbf{IMU + LiDAR + UWB Updates}} & 10.223 & 6.696 & 1.838 & 11.781 & 0.071 & 0.054 & 0.160 \\
\hline
\nohyphens{\textbf{IMU + LiDAR + UWB + Altimeter Updates}} & 9.491 & 6.623 & 0.435 & 11.574 & 0.073 & 0.055 & 0.139 \\
\hline
\nohyphens{\textbf{IMU + LiDAR + UWB + Altimeter + Hover (0.01) Updates}} & 1.583 & 7.465 & 0.232 & 7.631 & 0.056 & 0.087 & 0.082 \\
\hline
\nohyphens{\textbf{IMU + LiDAR + UWB + Altimeter + Hover (0.05) Updates}} & 5.595 & 7.972 & 0.606 & 8.437 & 0.074 & 0.092 & 0.128 \\
\hline
\nohyphens{\textbf{IMU + LiDAR + UWB + Altimeter + Hover (0.1) Updates}} & 1.665 & 7.428 & 0.462 & 7.614 & 0.077 & 0.088 & 0.105 \\
\hline
\nohyphens{\textbf{IMU + LiDAR + UWB + Altimeter + Velocity Updates}} & 0.637 & 0.741 & 0.122 & 0.865 & 0.067 & 0.063 & 0.116 \\
\hline
\nohyphens{\textbf{IMU + LiDAR + UWB + Altimeter + Hover + Velocity Updates}} & 0.924 & 0.840 & 0.336 & 1.043 & 0.069 & 0.087 & 0.098 \\
\hline
\end{tabularx}
\label{max_error}
\end{table*}

As shown the the analysis of the max errors, the 3D max error reduced by over 30\% (from 11.57 m to 7.61 m) just by including the hovering constraints.  

\section{Conclusions \& Future Work}
In this work, we systematically investigated the impact of incorporating hover constraints into a multi-sensor fusion framework for UAV-UGV cooperative localization in GNSS-denied environments. We introduced hover constraints into an incremental factor graph optimization framework by leveraging control command data from the UAV autopilot. Experimental results demonstrated that these hover constraints can improve state estimation accuracy when velocity updates are unavailable. This could be valueable in dark environments where computer vision based visual odometry is not possible. However, our analysis revealed that the benefits of hover constraints are diminished when direct velocity measurements are already included in the factor graph. Combining hover and velocity updates resulted in slightly higher RMS position errors than utilizing velocity updates alone, indicating potential measurement conflicts in the optimization scheme. This indicates that hover constraints could be most useful as a backup when velocity measurements are erroneous or unavailable. Our findings also indicate that while hover constraints improve position and attitude estimation, their effectiveness depends on the level of uncertainty assigned to the constraint. A balance between constraint tightness and realistic modeling of UAV hover dynamics is necessary to maximize performance improvements, since hovering is not zero-velocity.

\FloatBarrier

\bibliographystyle{IEEEtran}
\bibliography{sources}

\end{document}